\documentclass[journal]{IEEEtran}
\pdfoutput=1 

\usepackage{etoolbox}
\makeatletter
\patchcmd{\@makecaption}
  {\scshape}
  {}
  {}
  {}
\makeatother
\usepackage{epsfig}
\usepackage{graphicx}
\usepackage{amsmath}
\usepackage{amssymb}
\usepackage[nowarn]{glossaries}

\usepackage{booktabs}
\usepackage{relsize} % decrease font size for tables using relsize command
\usepackage{tabularx} % ragged (non-justified) table cells
\usepackage[newcommands]{ragged2e} % ragged table cells with hyphenated content
\usepackage[caption=false,font=scriptsize,justification=raggedright,singlelinecheck=true]{subfig} % subfigures
\usepackage{verbatim}  % multiline comments via \begin{comment}\end{comment}
\usepackage{url} % \url{my_url_here}.
\usepackage[multidot]{grffile} % allow dots in graphics filenames
\usepackage{paralist} %single-spaced lists via compactitem
\usepackage{cite} % smarter citations, e.g. makes [1][2][3] into [1]-[3]
\usepackage{textcomp} % \textdegree, use \textdegree\ to ensure space after

\usepackage[usenames,dvipsnames]{xcolor}
\usepackage[colorlinks,urlcolor=ForestGreen,citecolor=ForestGreen]{hyperref}
\usepackage{dblfloatfix}

\makeglossaries % on new projects run pdflatex then makeglossaries to make a .glo file, and add it to the repo (hack)
\usepackage{bm} % somewhat more comprehensive bolding with \bm

\DeclareMathSymbol{\Gamma}{\mathalpha}{letters}{"00}
\DeclareMathSymbol{\Delta}{\mathalpha}{letters}{"01}
\DeclareMathSymbol{\Theta}{\mathalpha}{letters}{"02}
\DeclareMathSymbol{\Lambda}{\mathalpha}{letters}{"03}
\DeclareMathSymbol{\Xi}{\mathalpha}{letters}{"04}
\DeclareMathSymbol{\Pi}{\mathalpha}{letters}{"05}
\DeclareMathSymbol{\Sigma}{\mathalpha}{letters}{"06}
\DeclareMathSymbol{\Upsilon}{\mathalpha}{letters}{"07}
\DeclareMathSymbol{\Phi}{\mathalpha}{letters}{"08}
\DeclareMathSymbol{\Psi}{\mathalpha}{letters}{"09}
\DeclareMathSymbol{\Omega}{\mathalpha}{letters}{"0A}

\newcommand{\vect}[1]{\bm{#1}}

\DeclareSymbolFont{EUr}{U}{eur}{m}{n}
\DeclareSymbolFont{EUb}{U}{eur}{b}{n}
\DeclareMathSymbol{\upGamma}{\mathord}{EUr}{"00}
\DeclareMathSymbol{\upDelta}{\mathord}{EUr}{"01}
\DeclareMathSymbol{\upTheta}{\mathord}{EUr}{"02}
\DeclareMathSymbol{\upLambda}{\mathord}{EUr}{"03}
\DeclareMathSymbol{\upXi}{\mathord}{EUr}{"04}
\DeclareMathSymbol{\upPi}{\mathord}{EUr}{"05}
\DeclareMathSymbol{\upSigma}{\mathord}{EUr}{"06}
\DeclareMathSymbol{\upUpsilon}{\mathord}{EUr}{"07}
\DeclareMathSymbol{\upPhi}{\mathord}{EUr}{"08}
\DeclareMathSymbol{\upPsi}{\mathord}{EUr}{"09}
\DeclareMathSymbol{\upOmega}{\mathord}{EUr}{"0A}
\DeclareMathSymbol{\upalpha}{\mathord}{EUr}{"0B}
\DeclareMathSymbol{\upbeta}{\mathord}{EUr}{"0C}
\DeclareMathSymbol{\upgamma}{\mathord}{EUr}{"0D}
\DeclareMathSymbol{\updelta}{\mathord}{EUr}{"0E}
\DeclareMathSymbol{\upepsilon}{\mathord}{EUr}{"0F}
\DeclareMathSymbol{\upzeta}{\mathord}{EUr}{"10}
\DeclareMathSymbol{\upeta}{\mathord}{EUr}{"11}
\DeclareMathSymbol{\uptheta}{\mathord}{EUr}{"12}
\DeclareMathSymbol{\upiota}{\mathord}{EUr}{"13}
\DeclareMathSymbol{\upkappa}{\mathord}{EUr}{"14}
\DeclareMathSymbol{\uplambda}{\mathord}{EUr}{"15}
\DeclareMathSymbol{\upmu}{\mathord}{EUr}{"16}
\DeclareMathSymbol{\upnu}{\mathord}{EUr}{"17}
\DeclareMathSymbol{\upxi}{\mathord}{EUr}{"18}
\DeclareMathSymbol{\uppi}{\mathord}{EUr}{"19}
\DeclareMathSymbol{\uprho}{\mathord}{EUr}{"1A}
\DeclareMathSymbol{\upsigma}{\mathord}{EUr}{"1B}
\DeclareMathSymbol{\uptau}{\mathord}{EUr}{"1C}
\DeclareMathSymbol{\upupsilon}{\mathord}{EUr}{"1D}
\DeclareMathSymbol{\upphi}{\mathord}{EUr}{"1E}
\DeclareMathSymbol{\upchi}{\mathord}{EUr}{"1F}
\DeclareMathSymbol{\uppsi}{\mathord}{EUr}{"20}
\DeclareMathSymbol{\upomega}{\mathord}{EUr}{"21}
\DeclareMathSymbol{\upvarepsilon}{\mathord}{EUr}{"22}
\DeclareMathSymbol{\upvartheta}{\mathord}{EUr}{"23}
\DeclareMathSymbol{\upvaromega}{\mathord}{EUr}{"24}
\DeclareMathSymbol{\upvarphi}{\mathord}{EUr}{"27}
\DeclareMathSymbol{\UpGamma}{\mathord}{EUb}{"00}
\DeclareMathSymbol{\UpDelta}{\mathord}{EUb}{"01}
\DeclareMathSymbol{\UpTheta}{\mathord}{EUb}{"02}
\DeclareMathSymbol{\UpLambda}{\mathord}{EUb}{"03}
\DeclareMathSymbol{\UpXi}{\mathord}{EUb}{"04}
\DeclareMathSymbol{\UpPi}{\mathord}{EUb}{"05}
\DeclareMathSymbol{\UpSigma}{\mathord}{EUb}{"06}
\DeclareMathSymbol{\UpUpsilon}{\mathord}{EUb}{"07}
\DeclareMathSymbol{\UpPhi}{\mathord}{EUb}{"08}
\DeclareMathSymbol{\UpPsi}{\mathord}{EUb}{"09}
\DeclareMathSymbol{\UpOmega}{\mathord}{EUb}{"0A}
\DeclareMathSymbol{\Upalpha}{\mathord}{EUb}{"0B}
\DeclareMathSymbol{\Upbeta}{\mathord}{EUb}{"0C}
\DeclareMathSymbol{\Upgamma}{\mathord}{EUb}{"0D}
\DeclareMathSymbol{\Updelta}{\mathord}{EUb}{"0E}
\DeclareMathSymbol{\Upepsilon}{\mathord}{EUb}{"0F}
\DeclareMathSymbol{\Upzeta}{\mathord}{EUb}{"10}
\DeclareMathSymbol{\Upeta}{\mathord}{EUb}{"11}
\DeclareMathSymbol{\Uptheta}{\mathord}{EUb}{"12}
\DeclareMathSymbol{\Upiota}{\mathord}{EUb}{"13}
\DeclareMathSymbol{\Upkappa}{\mathord}{EUb}{"14}
\DeclareMathSymbol{\Uplambda}{\mathord}{EUb}{"15}
\DeclareMathSymbol{\Upmu}{\mathord}{EUb}{"16}
\DeclareMathSymbol{\Upnu}{\mathord}{EUb}{"17}
\DeclareMathSymbol{\Upxi}{\mathord}{EUb}{"18}
\DeclareMathSymbol{\Uppi}{\mathord}{EUb}{"19}
\DeclareMathSymbol{\Uprho}{\mathord}{EUb}{"1A}
\DeclareMathSymbol{\Upsigma}{\mathord}{EUb}{"1B}
\DeclareMathSymbol{\Uptau}{\mathord}{EUb}{"1C}
\DeclareMathSymbol{\Upupsilon}{\mathord}{EUb}{"1D}
\DeclareMathSymbol{\Upphi}{\mathord}{EUb}{"1E}
\DeclareMathSymbol{\Upchi}{\mathord}{EUb}{"1F}
\DeclareMathSymbol{\Uppsi}{\mathord}{EUb}{"20}
\DeclareMathSymbol{\Upomega}{\mathord}{EUb}{"21}
\DeclareMathSymbol{\Upvarepsilon}{\mathord}{EUb}{"22}
\DeclareMathSymbol{\Upvartheta}{\mathord}{EUb}{"23}
\DeclareMathSymbol{\Upvaromega}{\mathord}{EUb}{"24}
\DeclareMathSymbol{\Upvarphi}{\mathord}{EUb}{"27}

\newcount\ppnum
\newcommand\ppnumber[1]{%
        \ppnum=#1\relax
        \ifnum\ppnum<0
                $-$%
                \ppnum=-\ppnum
        \fi
        \let\pptemp\empty
        \loop\ifnum\ppnum>999
                \count255=\ppnum
                \divide\ppnum by1000
                \count255=\numexpr \count255 - 1000*\ppnum \relax
                \edef\pptemp{,\!\ifnum\count255<100 0\ifnum\count255<10 0\fi\fi
                             \the\count255 \pptemp}%
        \repeat
        \the\ppnum
        \pptemp
}
\newacronym{PSI}{PSI}{Photonic Systems Integration}
\newacronym{ACFR}{ACFR}{Australian Centre for Field Robotics}
\newacronym{CRIS}{CRIS}{Centre for Robotics and Intelligent Systems}
\newacronym{ACRA}{ACRA}{the Australasian Conference on Robotics and Automation}
\newacronym{ACRV}{ACRV}{Australian Centre for Robotic Vision}
\newacronym{USyd}{USyd}{the University of Sydney}
\newacronym{UQ}{UQ}{the University of Queensland}
\newacronym{QUT}{QUT}{the Queensland University of Technology}
\newacronym{UCSD}{UCSD}{the University of California, San Diego}
\newacronym{ANU}{ANU}{Australia National University}
\newacronym{IMOS}{IMOS}{the Integrated Marine Observation System}
\newacronym{URI}{URI}{the University of Rhode Island}
\newacronym{WHOI}{WHOI}{Woods Hole Oceanographic Institution}

\newacronym{NSF}{NSF}{National Science Foundation}
\newacronym{LIEF}{LIEF}{Linkage Infrastructure, Equipment and Facilities}

\newacronym{ICCP}{ICCP}{the International Conference on Computational Photography}
\newacronym{CVPR}{CVPR}{Computer Vision and Pattern Recognition}
\newacronym{TIP}{TIP}{Transactions on Image Processing}
\newacronym{TSP}{TSP}{Transactions on Signal Processing}
\newacronym{JFR}{JFR}{the Journal of Field Robotics}
\newacronym{ISCAS}{ISCAS}{International Symposium on Circuits and Systems}
\newacronym{TOG}{TOG}{Transactions on Graphics}
\newacronym{ICRA}{ICRA}{International Conference on Robotics and Automation}
\newacronym{IROS}{IROS}{Intelligent Robots and Systems}
\newacronym{RA-L}{RA-L}{Robotics and Automation Letters}

\newacronym{AUV}{AUV}{autonomous underwater vehicle}
\newacronym{UAV}{UAV}{unmanned aerial vehicle}
\newacronym{USV}{USV}{unmanned surface vehicle}
\newacronym{UGV}{UGV}{unmanned ground vehicle}
\newacronym{GPS}{GPS}{global positioning system}
\newacronym{SLAM}{SLAM}{simultaneous localisation and mapping}
\newacronym{SfM}{SfM}{structure from motion}
\newacronym{AR}{AR}{augmented reality}
\newacronym{VR}{VR}{virtual reality}
\newacronym{MR}{MR}{mixed reality}
\newacronym{CNN}{CNN}{convolutional neural network}
\newacronym{IMU}{IMU}{inertial measurement unit}
\newacronym{TOF}{TOF}{time of flight}

\newacronym{MDSP}{MDSP}{multi-dimensional signal processing}
\newacronym{ROS}{ROS}{region of support}
\newacronym{DOF}{DOF}{degree-of-freedom}
\newacronym{RMS}{RMS}{root mean square}
\newacronym{SNR}{SNR}{signal-to-noise ratio}
\newacronym{CNR}{CNR}{contrast-to-noise ratio}
\newacronym{PCA}{PCA}{principal component analysis}
\newacronym{MSE}{MSE}{mean squared error}

\newacronym{FIR}{FIR}{finite impulse response}
\newacronym{IIR}{IIR}{infinite impulse response}
\newacronym{DFT}{DFT}{discrete Fourier transform}
\newacronym{FFT}{FFT}{fast Fourier transform}
\newacronym{PSNR}{PSNR}{peak signal-to-noise ratio}
\newacronym{FPGA}{FPGA}{field programmable gate array}
\newacronym{GPU}{GPU}{graphics processing unit}
\newacronym{ASIC}{ASIC}{application-specific integrated circuit}
\newacronym{BW}{BW}{bandwidth}

\newacronym{PSF}{PSF}{point spread function}
\newacronym{SPAD}{SPAD}{single-photon avalanche diode}
\newacronym{FOV}{FOV}{field of view}
\newacronym{BRDF}{BRDF}{bidirectional reflectance distribution function}
\newacronym{FWHM}{FWHM}{full width at half maximum}
\newacronym{LF}{LF}{light field}
\newacronym{2pp}{2pp}{two-plane parameterization}
\newacronym{MLA}{MLA}{microlens array}

\newacronym{RANSAC}{RANSAC}{random sampling and consensus}
\newacronym{DoG}{DoG}{difference of Gaussian}
\newacronym{SIFT}{SIFT}{scale invariant feature transform}

\DeclareMathOperator{\Filt2D}{Filt_{2D}}

\begin{document}

%%%%%%%%% TITLE
\title{LiFF: Light Field Features in Scale and Depth%
%\vspace{-1em}
}

% Donald, Bernd, Gordon
% List usyd and stan as my affiliations

\author{Donald G. Dansereau$^{1,2}$, Bernd Girod$^1$, and Gordon Wetzstein$^1$\\
$^1$Stanford University, $^2$The University of Sydney
%{\tt\small \{firstname.lastname\}@qut.edu.au}
%\and
%{\small\url{http://www.author.org/~second}}
}

%% omit the following lines up until the closing ``}''.
%% Additional authors and addresses can be added with ``\and'',
%% just like the second author.
%% To save space, use either the email address or home page, not both
%\and

% enable the \usepackage{authblk} to use this style of author block
%\author[1]{Donald G. Dansereau}
%\author[1]{Vincent Sitzmann}
%\author[1]{Jayant Thatte}
%\author[1]{Gordon Wetzstein}
%\affil[1]{Stanford University, Department of Electrical Engineering
%\authorcr  {\tt\small donald.dansereau@gmail.com}}
%%\affil[2]{University of California San Diego, Department of Electrical \& Computer Engineering
%%\authorcr {\tt\small gschuster@ucsd.edu, jeford@ucsd.edu}}
\maketitle
\thispagestyle{empty}  % the online instructions say no page numbers

\begin{abstract}
Feature detectors and descriptors are key low-level vision tools that many higher-level tasks build on. Unfortunately these fail in the presence of challenging light transport effects including partial occlusion, low contrast, and reflective or refractive surfaces. Building on spatio-angular imaging modalities offered by emerging light field cameras, we introduce a new and computationally efficient 4D light field feature detector and descriptor: LiFF. LiFF is scale invariant and utilizes the full 4D light field to detect features that are robust to changes in perspective. This is particularly useful for structure from motion (SfM) and other tasks that match features across viewpoints of a scene. We demonstrate significantly improved 3D reconstructions via SfM when using LiFF instead of the leading 2D or 4D features, and show that LiFF runs an order of magnitude faster than the leading 4D approach. Finally, LiFF inherently estimates depth for each feature, opening a path for future research in light field-based SfM.
\end{abstract}

%--------------------------------------------------------
\section{Introduction}
\label{Sec_intro}

Feature detection and matching are the basis for a broad range of tasks in computer vision.  Image registration, pose estimation, 3D reconstruction, place recognition, combinations of these, e.g.\ \gls{SfM} and \gls{SLAM}, along with a vast body of related tasks, rely directly on being able to identify and match features across images.  While these approaches work relatively robustly over a range of applications, some remain out of reach due to poor performance in challenging conditions. Even infrequent failures can be unacceptable, as in the case of autonomous driving.

\begin{figure}[t]
	\centering
	\includegraphics[width=\hsize]{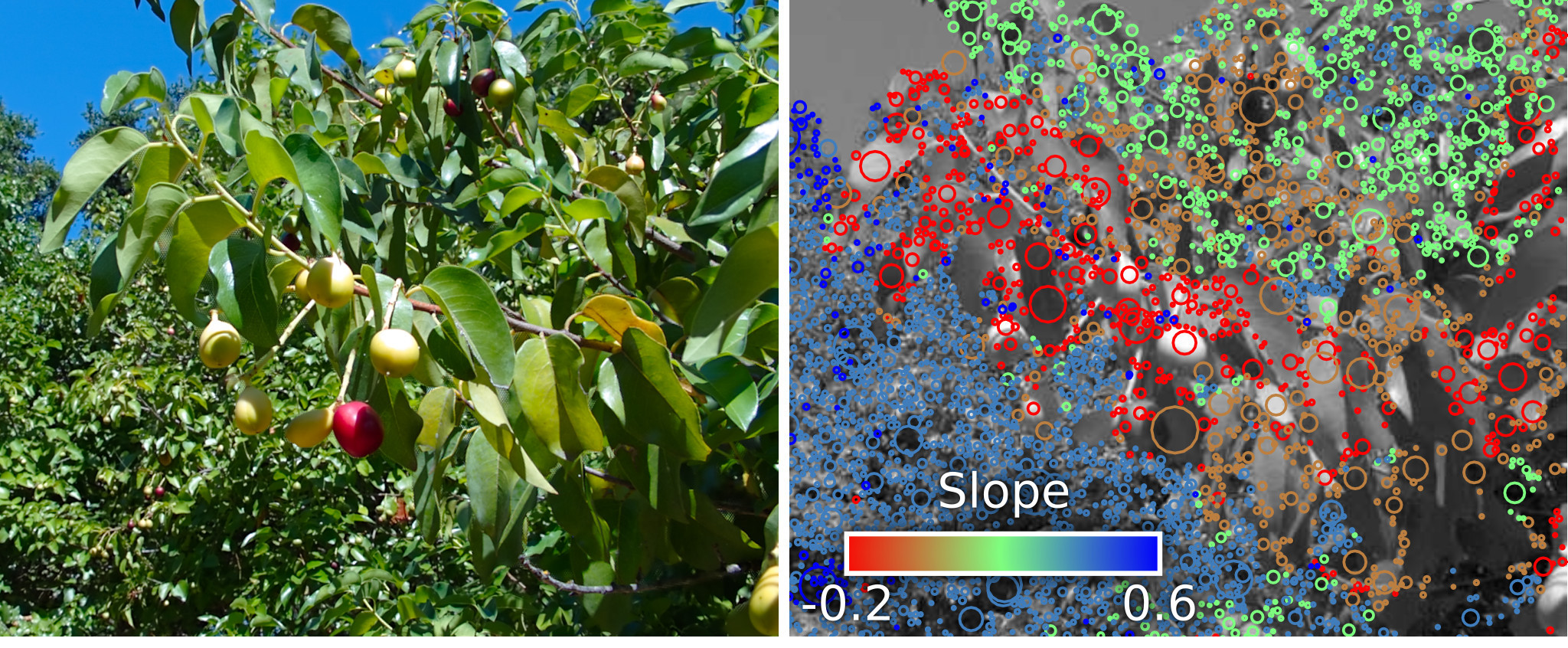}
	\caption{(left) One of five views of a scene that COLMAP's structure-from-motion (SfM) solution fails to reconstruct using SIFT, but successfully reconstructs using LiFF; (right) LiFF features have well-defined scale and depth, measured as light field slope, revealing the 3D structure of the scene~-- note we do not employ depth in the SfM solution. Code and dataset are at \url{http://dgd.vision/Tools/LiFF}, see the supplementary information for dataset details.
\vspace{-1em}
}
	\label{Fig_Teaser}
\end{figure}

State-of-the-art features fail in challenging conditions including self-similar, occlusion-rich, and non-Lambertian scenes, as well as in low-contrast scenarios including low light and scattering media.  For example, the high rate of self-similarity and occlusion in the scene in Fig.~\ref{Fig_Teaser} cause the COLMAP~\cite{schonberger2016structure} \gls{SfM} solution to fail.  There is also an inherent tradeoff between computational burden and robustness: given sufficient computation it may be possible to make sense of an outlier-rich set of features, but it is more desirable to begin with higher-quality features, reducing computational burden, probability of failure, power consumption, and latency.

\Gls{LF} imaging is an established tool in computer vision offering advantages in computational complexity and robustness to challenging scenarios~\cite{wanner2013reconstructing, dong2013plenoptic, mitra2014framework, dansereau2015linear_paper, srinivasan2017light}.  This is due both to a more favourable \gls{SNR} / depth of field tradeoff than for conventional cameras, and to the rich depth, occlusion, and native non-Lambertian surface capture inherently supported by \glspl{LF}.

In this work we propose to detect and describe blobs directly from 4D \glspl{LF} to deliver more informative features compared with the leading 2D and 4D alternatives. Just as the \gls{SIFT} detects blobs with well-defined scale, the proposed light field feature (LiFF) identifies blobs with both well-defined scale and well-defined depth in the scene. Structures that change their appearance with viewpoint, for example those refracted through or reflected off curved surfaces, and those formed by occluding edges, will not satisfy these criteria. At the same time, well-defined features that are partially occluded are not normally detected by 2D methods, but can be detected by LiFF via focusing around partial occluders.

Ultimately LiFF features result in fewer mis-registrations, more robust behaviour, and more complete 3D models than the leading 2D and 4D methods, allowing operation over a broader range of conditions.   Following recent work comparing hand-crafted and learned features~\cite{schonberger2017comparative}, we evaluate LiFF in terms of both low-level detections and the higher-level task of 3D point cloud reconstruction via \gls{SfM}.

LiFF features have applicability where challenging conditions arise, including autonomous driving, delivery drones, surveillance, and infrastructure monitoring, in which weather and low light commonly complicate vision.  It also opens a range of applications in which feature-based methods are not presently employed due to their poor rate of success, including medical imagery, industrial sites with poor visibility such as mines, and in underwater systems.

The key contributions of this work are:
\begin{compactitem}
\item We describe LiFF, a novel feature detector and descriptor that is less computationally expensive than leading 4D methods and natively delivers depth information;
\item We demonstrate that LiFF yields superior detection rates compared with competing 2D and 4D methods in low-SNR scenarios; and
\item We show that LiFF extends the range of conditions under which SfM can work reliably, outperforming SIFT in reconstruction performance.
\end{compactitem}

To evaluate LiFF we collected a large multi-view \gls{LF} dataset containing over 4000 \glspl{LF} of over 800 scenes.  This is the first large dataset of its kind, with previous examples limited to a single \gls{LF} of each scene~\cite{srinivasan2017learning}. It is our hope that LiFF and the accompanying dataset will stimulate a broad range of research in feature detection, registration, interpolation, SfM, and SLAM.

%------------------------------
\section{Related Work}
\label{sec:related}

\paragraph{Feature Detection and Matching}

2D feature detectors such as SIFT~\cite{lowe2004distinctive}, SURF~\cite{bay2008surf}, FAST~\cite{rosten2006machine}, and ORB~\cite{rublee2011orb}, are instrumental in many computer vision algorithms, including \gls{SfM}, \gls{SLAM}, disparity estimation, and tracking. Many of these applications rely on matching features between different viewpoints of the same scene. Unfortunately, this matching is often unreliable, because similar spatial structures can occur several times in the same scene, and view-dependent effects such as partial occlusion and specularity makes features look different from different perspectives. To reliably match features, additional geometric constraints have to be imposed, for example via bundle adjustment, but this is computationally expensive, severely affecting runtime, required memory, and power.

3D feature detection from RGB-D images can be more robust than 2D feature detection, as demonstrated in the context of object detection and segmentation~\cite{gupta2014learning} as well as \gls{SLAM}~\cite{gao2015robust}. Rather than working with RGB-D data, 3D feature detectors can also operate directly on point clouds~\cite{gumhold2001feature, zhong2009intrinsic, steder2010narf} while providing similar benefits. However, point clouds are usually not available in conventional imaging systems and RGB-D data does not generally handle partial occlusion and other view-dependent effects. 

\Glspl{LF} inherently capture a structured 4D representation that includes view-dependent effects, partial occlusions, and depth. A number of existing works touch upon exploiting these characteristics for feature detection and description.  Ghasemi et al.~\cite{ghasemi2014scale} exploit depth information in the \gls{LF} to build a global, scale-invariant descriptor useful for scene classification, though they do not address the localized features required for 3D reconstruction.  Tosic et al.~\cite{tosic20143d} employ \gls{LF} scale and depth to derive an edge-sensitive feature detector. Our focus is on blob detection rather than edge detection since it is much easier to uniquely match blobs across viewpoints, making it appropriate for a larger set of tasks including 3D reconstruction. 

The leading method for extracting features from \glspl{LF} is to run a 2D detector across subimages, then consolidate the detected features by imposing consistency with epipolar geometry.  For example, Teixeira et al.~\cite{teixeira2017epipolar} propose feature detection that repeats SIFT on 2D subimages then consolidates across 2D epipolar slices.  In exploring \gls{LF} \gls{SfM}, Johannsen et al.~\cite{johannsen2015linear} extract SIFT features across subimages then consolidate them using 4D \gls{LF} geometry.  Zhang et al.~\cite{zhang2017ray} demonstrate that line and plane correspondence can be employed for \gls{LF}-based \gls{SfM}, by detecting 2D line segments in subimages then applying a higher-order consolidation step. Finally, Maeno et al.~\cite{maeno2013light} and Xu et al.~\cite{xu2015transcut} detect refractive objects by following 2D features through the \gls{LF} using optical flow. They then enforce 4D epipolar geometry to detect refracted features.

While these approaches differ in their details they are all fundamentally limited by the performance of the 2D detector they build upon.  We refer to these as repeated 2D detectors, and make direct comparison to repeated SIFT in this work.  We show that LiFF shows significantly higher performance over repeated 2D methods by virtue of simultaneously considering all subimages in detecting and describing features.  Because repeated 2D detectors are less direct in their approach, they present more parameters requiring tuning, making them more difficult to deploy.  Finally, repeating SIFT across viewpoints is a highly redundant operation, and we will show that LiFF has significantly lower computational complexity.

\paragraph{Light Field Imaging}
An \gls{LF}~\cite{levoy1996light, gortler1996lumigraph} contains 4D spatio-angular information about the light in a scene, and can be recorded with a camera array~\cite{wilburn2005high}, or a sensor equipped with a lenslet array~\cite{adelson1992single, ng2005light, dansereau2017wide} or a coded mask~\cite{veeraraghavan2007dappled, marwah2013compressive}.  See~\cite{ihrke2016principles, wetzstein2011computational} for detailed overviews of \gls{LF} imaging. To date, \gls{LF} image processing has been applied to a variety of applications including image-based rendering~\cite{levoy1996light, davis2012unstructured, levin2010linear}, post-capture image refocus~\cite{ng2005fourier, georgiev2008unified}, \gls{SfM}~\cite{johannsen2015linear}, lens aberration correction~\cite{hanrahan2006digital}, spatial~\cite{bishop2009light} and temporal~\cite{wang2017light} super-resolution, video stabilization~\cite{smith2009light}, motion deblurring~\cite{srinivasan2017light}, and depth imaging~\cite{wanner2012globally, wanner2013globally, tao2015depth, liang2015light, tian2017depth}. In this work, we explore robust \gls{LF} feature detection and matching for improving applications in reconstruction including \gls{SfM}. 

\paragraph{Conventions}
In this work we consider the two-plane-parameterized \gls{LF} $L(s,t,u,v)$ with $N_s \times N_t$ views of $N_u \times N_v$ pixels each~\cite{levoy1996light, dansereau2015linear_paper}.  A point in 3D space appears in the LF as a plane with slope inversely proportional to the point's depth~\cite{bolles1987epipolar, adelson1992single, ihrke2016principles}.   Working with sampled \glspl{LF} introduces unknown scaling factors between slope and depth, which can either be tolerated or calibrated away.  In the following we refer to slope with the understanding that it can be mapped to depth via camera calibration~\cite{dansereau2013decoding, bok2014geometric, xu2015camera}.

% of the form
%\begin{equation}
%\begin{bmatrix}
%u\\v
%\end{bmatrix}
%= 
%-{\frac{D}{P_z}}
%\begin{bmatrix}
%s - P_x \\
%t - P_y
%\end{bmatrix}
%\label{eq_PtPlane}
%\end{equation}
%where the plane slope $-D/P_z$ is inversely proportional to the point's depth in the scene $P_z$~\cite{bolles1987epipolar, adelson1992single, ihrke2016principles}, allowing depth to be inferred from slope.   Working with sampled \glspl{LF} introduces unknown scaling factors between slope and depth.  Many applications employ such sampled slopes directly, while others use camera calibration to map slope to metric depth~\cite{dansereau2013decoding}. In the following we work with slope with the understanding that it maps directly to depth.

%------------------------------
\section{Light Field Feature Detection}

We begin our development with the well-known SIFT feature detector and extend it to 4D \glspl{LF}. We begin with SIFT because of its dominance in reconstruction applications~\cite{schonberger2017comparative}.  Our key insight is that while SIFT locates blobs with well-defined scales and locations in the 2D image plane, \glspl{LF} offer the ability to identify blobs with well-defined scales and locations \emph{in 3D space}.  This offers numerous advantages including rejection of undesired spurious features at occlusion boundaries, detection of desired but partially occluded features, and inherent depth estimation.

SIFT identifies blobs by searching for extrema in a 3D scale space constructed as a \gls{DoG} stack.  The \gls{DoG} is built by convolving with a set of Gaussian filters covering a range of scales, then taking the difference between adjacent scales, as in
\begin{align}
L(x,y,\sigma) &= G(x,y,\sigma) * I(x,y),\label{eq_SIFT_convolution}\\
D(x,y,\sigma_i) &= L(x,y,\sigma_{i+1}) - L(x,y,\sigma_i), \label{eq_DOG}
\end{align}
where $G(x,y,\sigma)$ is a Gaussian filter at scale $\sigma$, and the \gls{DoG} is computed over a range of scales $\sigma_i, 1 \le i \le N$ with constant multiplicative factor $k$ such that $\sigma_{i+1} = k \sigma_{i}$.

The convolutions \eqref{eq_SIFT_convolution} represent the bulk of the computational cost of SIFT.  Significant savings can be had by applying larger-scaled convolutions on downsampled versions of the input image~\cite{lowe2004distinctive}.  Nevertheless, a good approximation of the cost of this approach is to understand it as a set of $N$ 2D filtering operations, which we denote $N \times \Filt2D$.

Following extrema detection, SIFT proceeds through steps for sub-pixel-accurate feature location, rejection of edge features that can trigger the blob detection process, and estimation of dominant orientation allowing rotation invariance.  Finally, an image descriptor is constructed from histograms of edge orientations.  LiFF will differ from these steps only in the detection and descriptor stages.

%%----
%\subsection{Repeating SIFT over 4D LFs}

%A common approach to \gls{LF} feature detection is repeating SIFT across all \gls{LF} subimages, then applying a consistency check to reject inconsistent features~\cite{teixeira2017epipolar, johannsen2015linear}.  The complexity of using this approach with SIFT is at least the cost of the DoG operations applied over the subimages, i.e.~$N_s \times N_t \times \times N \times \Filt2D$.  Note that this ignores the cost of consolidating observations across views, which varies by implementation and can be substantial.  In this work we will show repeated SIFT to offer inferior performance compared with the natively 4D LiFF.

%----
\subsection{Searching Scale and Slope}

To generalize SIFT to the \gls{LF} we first propose a much more computationally expensive approach that offers enhanced performance compared with SIFT and repeated SIFT.  We then show how this approach can be implemented much more efficiently with numerically identical results, making it more computationally efficient than repeating SIFT across the \gls{LF}.

Jointly searching across scale and 3D position can be accomplished as a direct extension of SIFT's DoG space.  We first rewrite each scale of the \gls{DoG}~\eqref{eq_DOG} as a single convolution, applied in the $u$ and $v$ dimensions
\begin{gather}
	H_{\sigma}(u,v,\sigma) = G(u,v,\sigma_{i+1}) - G(u,v,\sigma),\label{eq_DoG} \\
	D_{2D}(u,v,\sigma) = H_{\sigma}(u,v,\sigma) * I(u,v).
\end{gather}
The filter $H_{\sigma}$ finds blobs in \gls{LF} subimages at the scale $\sigma$.  We augment this with depth selectivity using a frequency-planar filter $H_{\lambda}$.  The frequency-planar filter selects for a specific depth in the \gls{LF}, and can be constructed in a number of ways in the frequency or spatial domains~\cite{ng2005fourier, dansereau2015linear_paper}.  For this work we consider the direct spatial-domain implementation
\begin{gather}
	H_{\lambda}(s,t,u,v,\lambda) = 
	\begin{cases}
		1,	& u = \lambda s,\,\, v = \lambda t,\\
		0,  & \text{otherwise.}
	\end{cases}
	\label{eq_Planar}
\end{gather}
We combine \eqref{eq_DoG} and \eqref{eq_Planar} to yield a filter that is simultaneously selective in scale and slope:
\begin{equation}
	H(\vect{\phi}, \sigma,\lambda) = H_\sigma( u,v, \sigma ) * H_\lambda( \vect{\phi}, \lambda ),\label{eq_4DScaleSlope}
\end{equation}
where $\vect{\phi} = [s,t,u,v]$ gathers the \gls{LF} indices. We apply the filter $H$ over $N$ scales $\sigma$ and $M$ slopes $\lambda$:
\begin{equation}
	D_{6D}(\vect{\phi},\sigma,\lambda) = H(\vect{\phi}, \sigma,\lambda) * L(\vect{\phi}).
	\label{eq_Full4DConv}
\end{equation}
$D_{6D}$ is highly redundant in that each subimage contains virtually the same information, and so when searching for local extrema we restrict our attention to the central view in $s,t$ yielding the 4D search space $D(u,v,\sigma,\lambda)$.

Identifying local extrema in $D$ is a straightforward extension of the 3D approach used in SIFT, yielding feature coordinates $[u,v,\sigma,\lambda]$. It is important to jointly search the scale-slope space in order to identify those features with both distinct scale and slope. This is a key distinction between LiFF and repeating SIFT over the LF or a focal stack.

%----
\subsection{Simplification using Focal Stack}

The method so far is extremely computationally expensive.  The 4D convolution~\eqref{eq_Full4DConv} is repeated over $N$ scales and $M$ slopes.  The key insight in simplifying~\eqref{eq_Full4DConv} is exploiting the linear separability of $H_\sigma$ and $H_\lambda$ seen in \eqref{eq_4DScaleSlope}.  The fact that we employ only the central view of $D$ allows the slope selectivity step to be computed only over that subset, collapsing the 4D \gls{LF} into a 3D focal stack:
\begin{gather}
F(u,v,\lambda) = \sum_{s,t} L(s,t, u-\lambda s, v-\lambda t),\label{eq_FocalStack}\\
D(u,v,\sigma,\lambda) = H_{\sigma}(u,v,\sigma) * F(u,v,\lambda).\label{eq_Efficient}
\end{gather}
i.e. we compute a focal stack $F$ over $M$ slopes, then apply a \gls{DoG} filter over $N$ scales for each slope. Finally, we search the joint space $D$ for extrema.  This process yields numerically identical results to building the full 6D scale-slope space~\eqref{eq_Full4DConv}, but at a fraction of the computational cost.

A few efficient methods for computing the focal stack $F$ have been proposed~\cite{perez2014fast, marichal2011fast}.  These generally find at minimum as many layers as there are samples in $s$ or $t$.  Feature detection may not require so many layers, and so we proceed with the more straightforward approach of shifting and summing \gls{LF} subimages~\eqref{eq_FocalStack}, with the understanding that computational savings may be possible for large stack depths.  The cost of this focal stack is $M \times N_s \times N_t \times N_u \times N_v$.

Computing the \gls{DoG} from each focal stack image $F$ is identical to the first steps of conventional SIFT, and can benefit from the same downsampling optimization~\cite{lowe2004distinctive}.  We approximate the complexity as $M$ times the cost of conventional SIFT, $M \times N \times \Filt2D$.  For practical scenarios this will overshadow the cost of computing the focal stack.

%----
\subsection{Feature Descriptor}

As with SIFT, for each feature $[u,v,\sigma,\lambda]$ we construct a histogram of edge orientations.  The key difference with the LiFF descriptor is that it is computed at a specific depth in the scene corresponding to the detected slope $\lambda$.  Each descriptor is thus constructed from the appropriate stack slice $F(u,v,\lambda)$.  The key advantage is selectivity against interfering objects at different depths including partial occluders and reflections off glossy surfaces.

%----
\subsection{Complexity}
\label{Sect_Complexity}

From the above we note that for $M$ slopes, LiFF is $M$ times slower than SIFT applied to a single subimage.  However, we will show that LiFF delivers more informative features than SIFT by virtue of its higher detection rates in low contrast and noise, better resilience to spurious features and partial occlusions, and inherent estimation of slope.

A common approach to \gls{LF} feature detection is repeating SIFT across subimages, then applying a consistency check to reject spurious detections~\cite{teixeira2017epipolar, johannsen2015linear}.  The complexity of this approach is at least the cost of the DoG operations applied over the subimages, i.e.~$N_s \times N_t \times N \times \Filt2D$.  Note that this ignores the cost of consolidating observations across views, which varies by implementation and can be substantial.  We will show that repeated SIFT also offers inferior performance compared with the natively 4D LiFF.

Comparing complexity, we see that LiFF is at least $N_s N_t/M$ times faster than repeated SIFT.  In a typical scenario using Lytro Illum-captured LFs with $11 \times 11$ views, and applying LiFF over $M=11$ slopes, LiFF will be about 11 times slower than SIFT, but 11 times faster than repeated SIFT.  For larger \glspl{LF}, e.g.~Stanford gantry-collected \glspl{LF}\footnote{\url{http://lightfields.stanford.edu}} with $17 \times 17$ views, the speed increase is larger, 26 times, assuming the same slope count.  

%----
\subsection{Parameters}

LiFF has the same parameters as SIFT: a list of scales at which to compute the DoG, a peak detection threshold, and an edge rejection threshold. The descriptor parameters are also the same, including the area over which to collect edge histograms, numbers of bins, and so on.  The only additional parameter for LiFF is a list of slopes over which to compute the focal stack.  A good rule of thumb for lenslet-based cameras is to consider slopes between -1 and 1, with as many slopes as there are samples in $N_s$ or $N_t$.  Larger slope counts increase compute time without improving performance, while smaller slope counts can miss features at specific depths in the scene.

%--------------------------------------------------------
\section{Evaluation}

\begin{figure*}
	\centering
	\includegraphics[width=0.7\hsize]{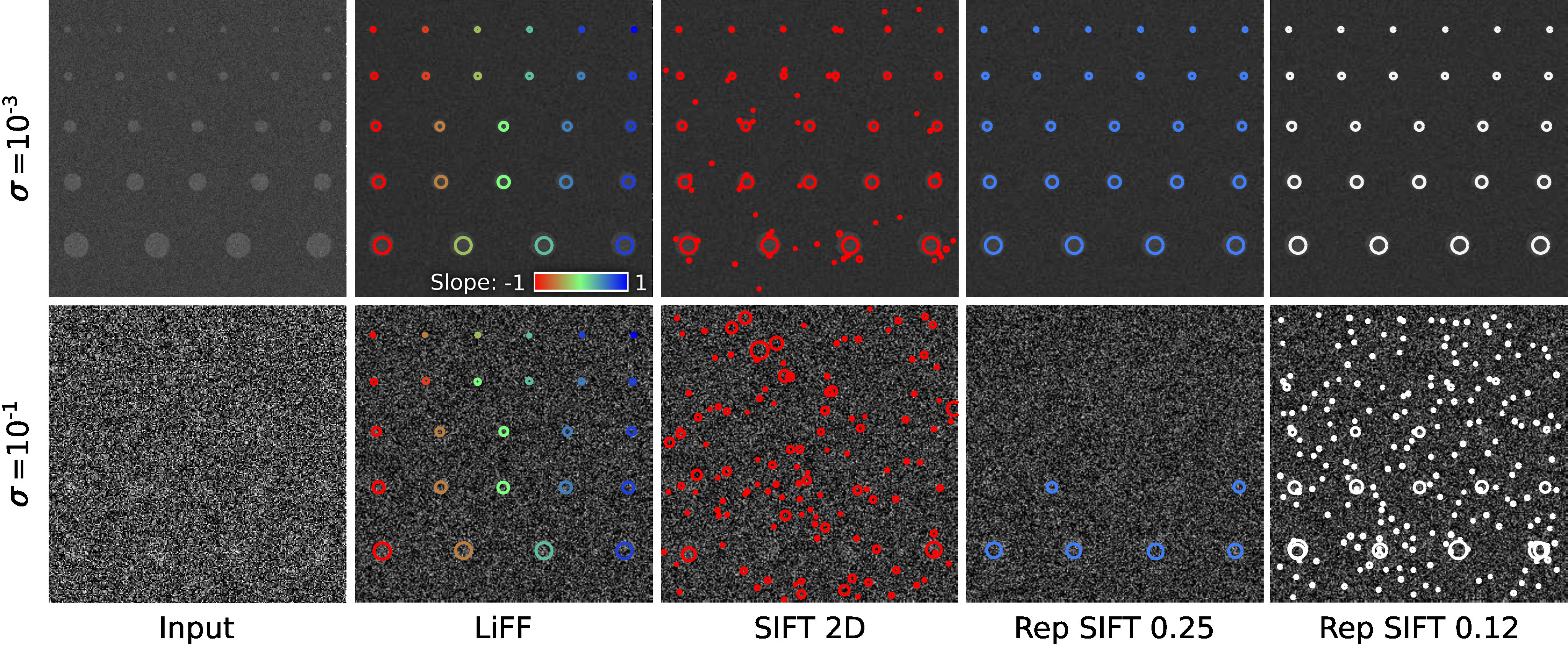}\hfil
	\caption[]{A set of disks at varying scales and depths, presented at two noise levels $\sigma$. At the lower noise level (top row), all methods operate reasonably well; while SIFT shows some spurious detections, repeated SIFT is able to reject these by imposing consistency between views; In higher noise (bottom row) LiFF's performance is ideal including reasonable slope estimates, but SIFT misses some features and has spurious detections; repeated SIFT with threshold 0.25 rejects the spurious features but cannot locate those missed in the individual views; and repeated SIFT with a lower threshold admits more spurious detections while still missing some true positives.%
%\vspace{-1em}
}
\label{Fig_SynthSnrExample}
\end{figure*}

\paragraph{LiFF Implementation} Our implementation of LiFF is in C, compiled into MEX files that we call from MATLAB.  For testing purposes, we load light fields and convert to grayscale in MATLAB, but the feature detection and extraction process is entirely in C. Our focal stack implementation uses the shift-and-sum method with nearest-neighbour interpolation, and includes a normalization step which prevents darkening near the edges of the \gls{LF}.

\paragraph{Repeated SIFT Implementation}
To compare LiFF with repeated SIFT, we called the VLFeat C implementation of SIFT v0.9.21, and in MATLAB implemented a consolidation process that enforces consistency between subimages.  A variety of approaches have been suggested~\cite{teixeira2017epipolar, johannsen2015linear, maeno2013light, xu2015transcut}.  Our goal for SfM testing is not speed, but approaching the upper bound of performance.  We therefore employ an exhaustive search starting at each detected 2D feature across all subimages.  For each feature we identify matching detections in all other subimages based on a set of criteria including scale, orientation, feature descriptor, and maximum deviation from a best-fit plane. When evaluating speed we omit the time taken for this consolidation process.

A key parameter of any repeated 2D detector is the number of subimages in which a feature must be identified before being considered a detection.  In the following we test across different thresholds, and identify the method accordingly, e.g.~repeated SIFT 0.5 requires that at least half of the subimages contain a detected feature.

Our repeated SIFT implementation is not very computationally efficient. However we believe its performance is revealing of a broad class of repeated and consolidated 2D features.

%---
\subsection{Speed}
We compared the speed of our LiFF implementation with the SIFT implementation in VLFeat.  All tests were run on an Intel i7-8700 at 3.20~GHz.  The test included both feature detection and descriptor extraction, and was run on scenes with similar feature counts for SIFT and LiFF.  On Illum-captured LFs with $11 \times 11 \times 541 \times 376$ samples, we found LiFF took on average 2.88~sec, while repeating SIFT across subimages took on average 53.1~sec, excluding time to consolidate observations, which was considerable.

Overall the speed increase moving from repeated SIFT to LiFF with our implementation is measured as $18 \times$, which agrees well with the anticipated speed gain. Further speed improvements should be possible: as with SIFT, LiFF is amenable to optimization via parallelization, implementation on GPU, etc.

%----
\subsection{Noise Performance}

Repeated SIFT is fundamentally constrained by the performance of the 2D method it builds on.  To demonstrate this we synthesized a set of scenes with known good feature locations, and introduced varying levels of noise to observe feature performance.

In one set of experiments, depicted in Fig.~\ref{Fig_SynthSnrExample}, the input consists of 26 disks at varying scales and at depths corresponding to slopes between -1 and 1.  The \gls{LF} has dimensions $9 \times 9 \times 256 \times 256$ and a signal contrast of 0.1.  We introduced moderate noise with variance $10^{-3}$ (top), and strong noise with variance $10^{-1}$ (bottom).

We ran SIFT operating on the central subimage of the LF, repeated SIFT with minimum subimage agreements of 0.25 and 0.12, and LiFF.  The common parameters of peak threshold, edge detection threshold, and scale range were identical for all methods.  LiFF was operated over 9 slopes between -1 and 1.

As seen in Fig.~\ref{Fig_SynthSnrExample}, LiFF successfully detects all 26 disks in both moderate and high noise, as well as providing slope estimates even in high noise.  SIFT suffers from spurious detections in moderate noise, and both missed and spurious detections in high noise.  Repeated SIFT successfully rejects spurious detections in low noise, but either misses detections or both misses detections and admits spurious features in high noise, depending on its threshold.

\begin{figure}[b!]
	\centering
	\subfloat[]{\includegraphics[height=0.4\hsize]{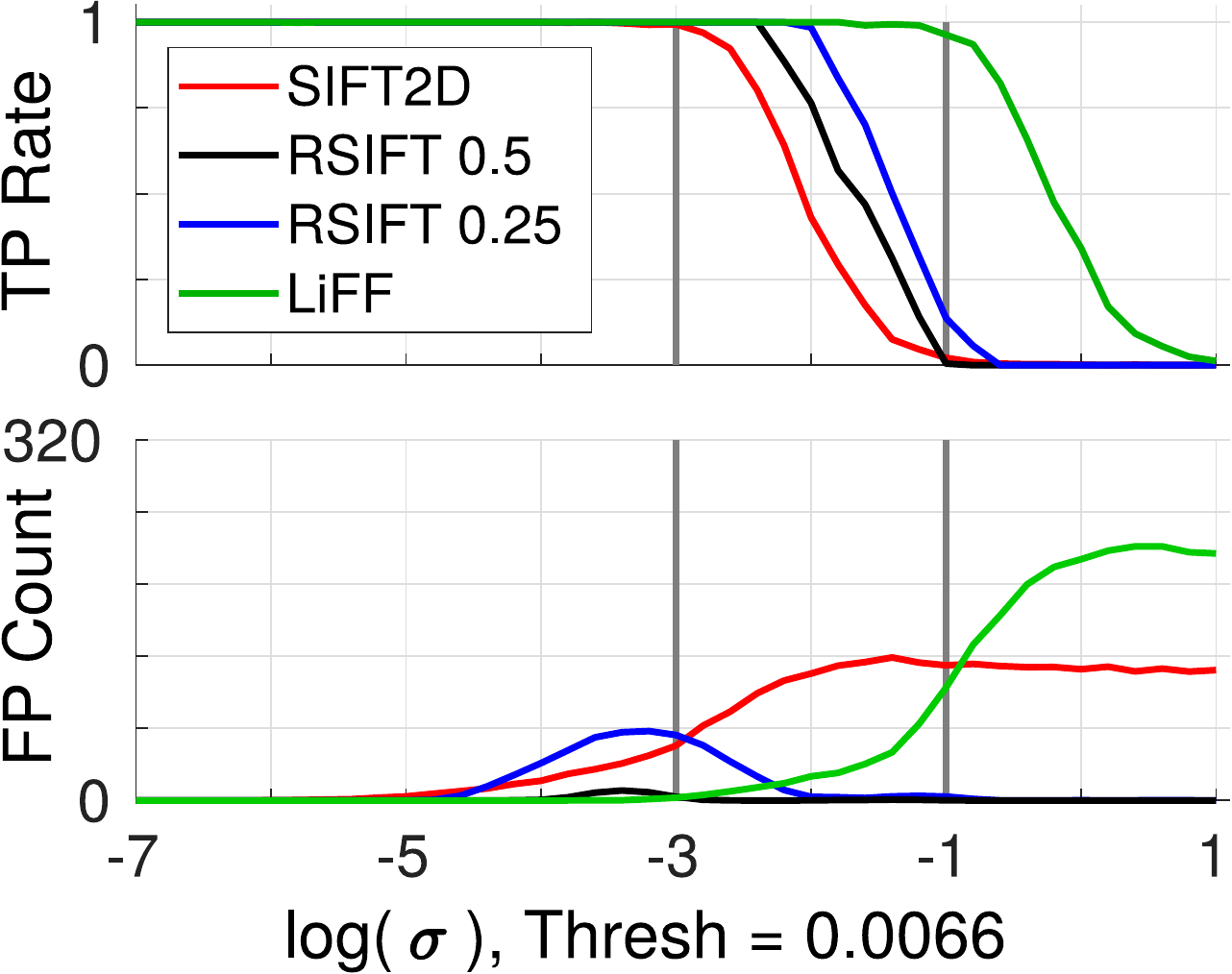}\label{Fig_SNRSweep_T1}}\hfil
	\subfloat[]{\includegraphics[height=0.4\hsize]{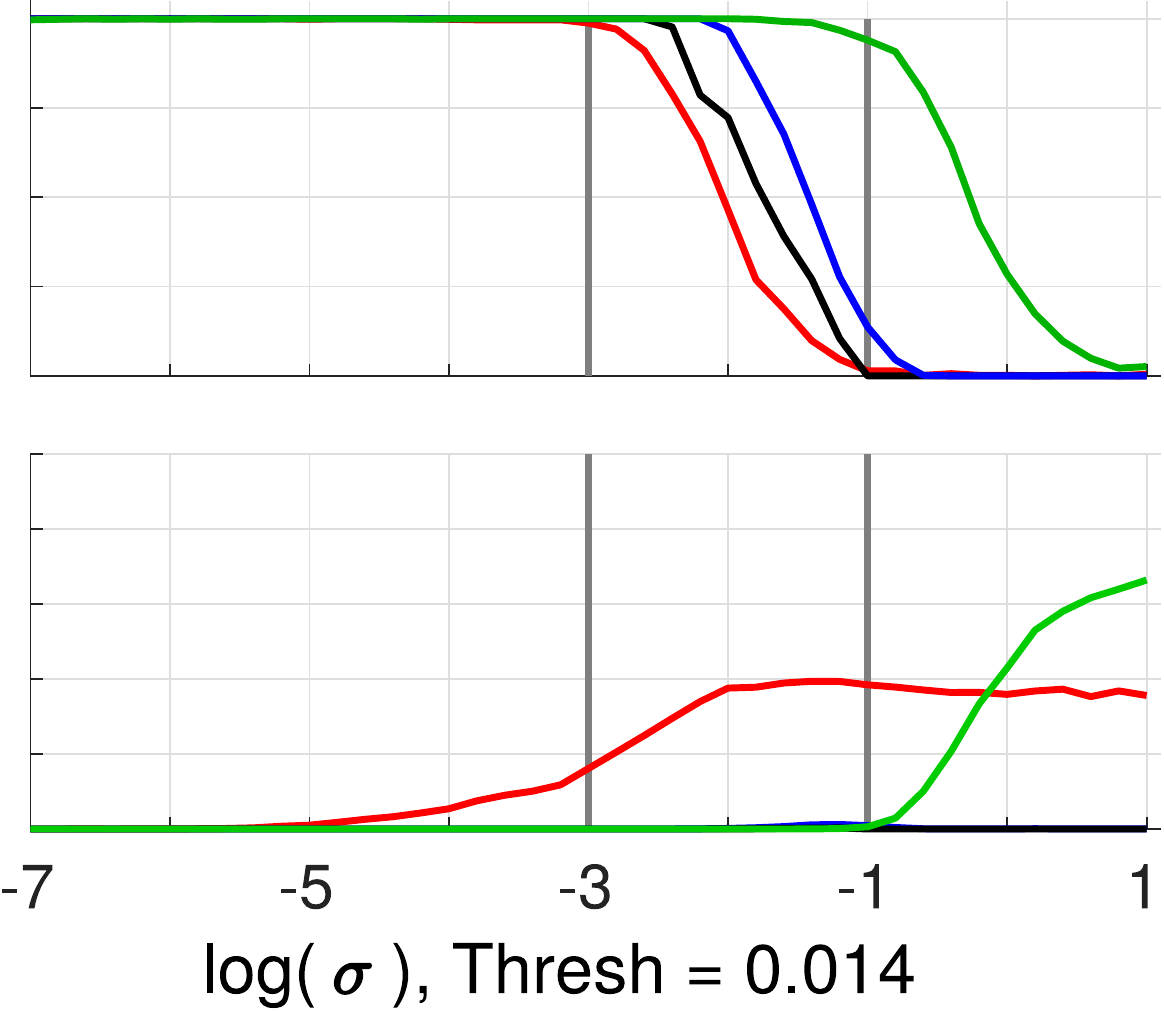}\label{Fig_SNRSweep_T2}}\\
	\subfloat[]{\includegraphics[height=0.39\hsize]{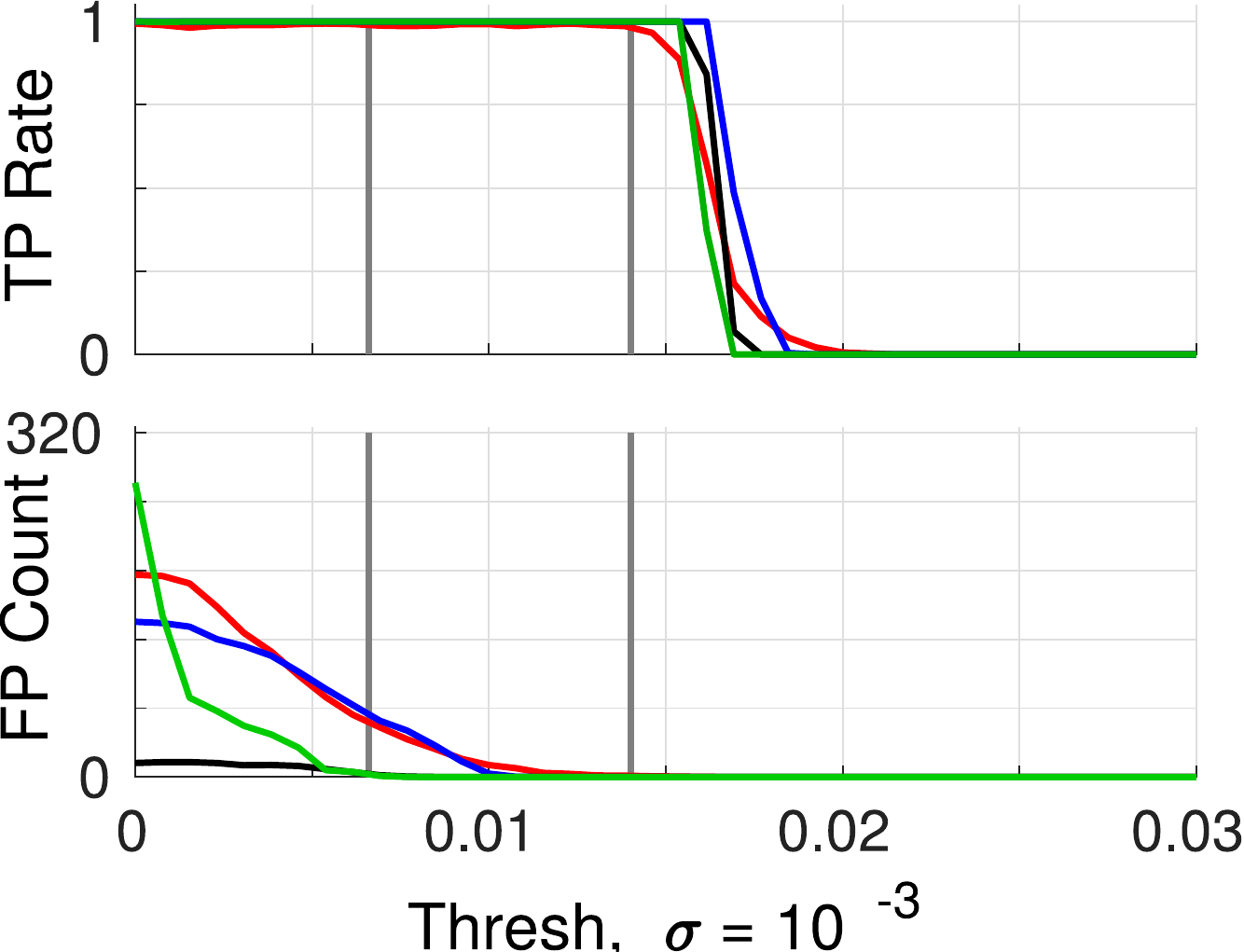}\label{Fig_ThreshSweep_N3}}\hfil
	\subfloat[]{\includegraphics[height=0.39\hsize]{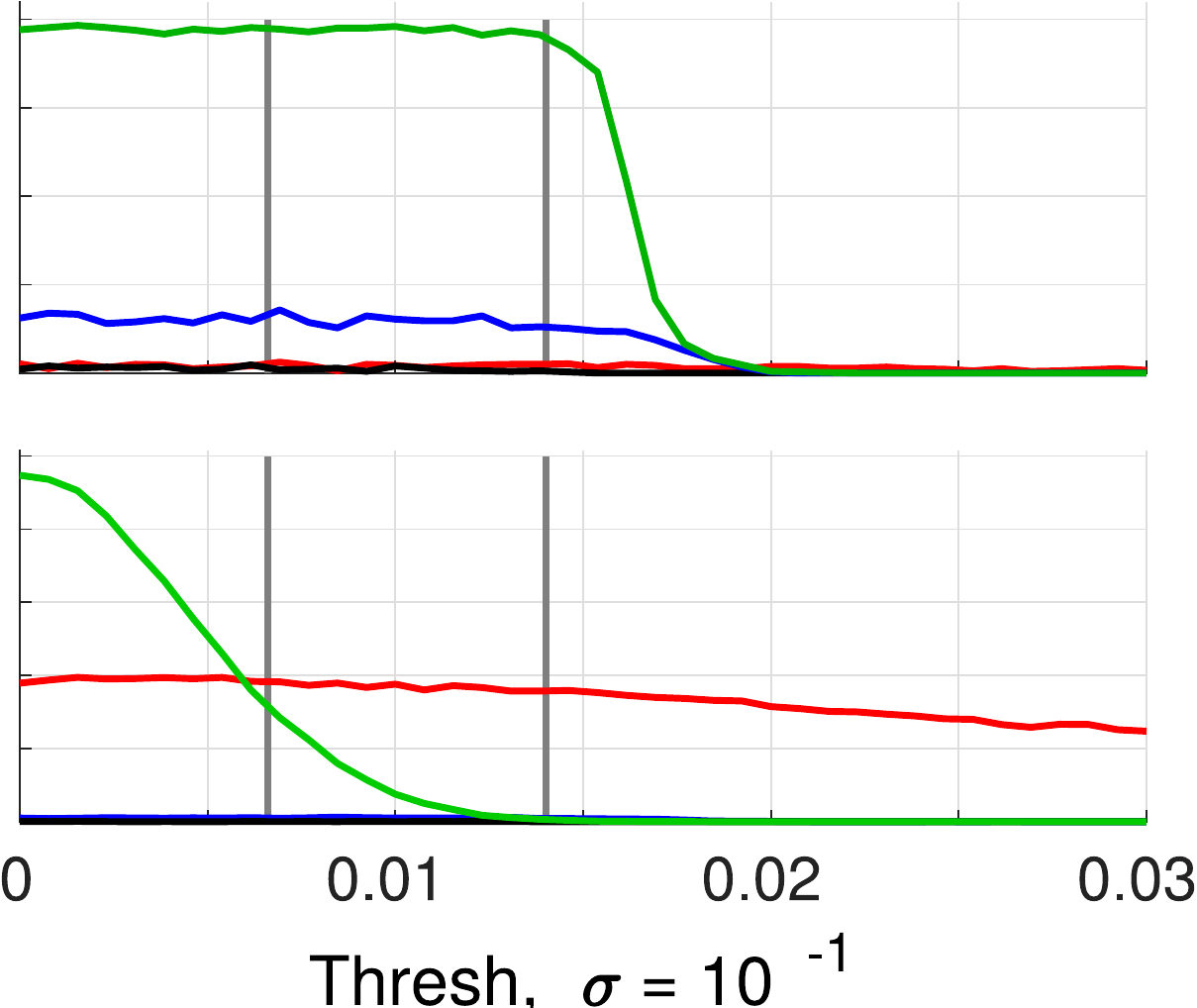}\label{Fig_ThreshSweep_N1}}
	\caption[]{Noise performance: (a,b) Sweeping noise level $\sigma$ for fixed detection thresholds, LiFF has the best true positive (TP) rate for noisy imagery, though like SIFT suffers from a high false positive (FP) count; (c) Sweeping detection threshold, the methods show similar performance in moderate noise, while (d)~LiFF delivers a much higher TP rate and zero FP rate in high noise for appropriately set threshold.  Overall, LiFF matches or outperforms both SIFT and repeated SIFT.
%\vspace{-1em}
}
\label{Fig_NoisePerformance}
\end{figure}

To better expose the behaviours of these methods we ran a set of experiments on the same scene with varying noise levels and peak detection thresholds, measuring true positive (TP) rate over the 26 disks, and false positive (FP) count.  Each experiment was repeated 25 times, with the mean results shown in Fig.~\ref{Fig_NoisePerformance}.  The top row depicts two detection thresholds (highlighted as vertical bars on the bottom row), with noise variances $\sigma$ swept between $10^{-7}$ and $10^{1}$.  The TP rate shows that LiFF correctly detects features in more than an order of magnitude higher noise than the other methods.  At high noise levels LiFF and SIFT both suffer from high FP counts, though this is somewhat ameliorated for LiFF by setting a higher peak detection threshold.

The bottom row of Fig.~\ref{Fig_NoisePerformance} depicts two noise levels, $\sigma = 10^{-3}$ and $10^{-1}$ (highlighted as vertical bars in the top row), for varying peak detection thresholds.  In moderate noise (left) all methods perform similarly across a range of threshold values.  In high noise (right), only LiFF delivers a good TP rate, and a nil FP count for a sufficiently large detection threshold.

From these experiments we conclude that LiFF offers enhanced performance in noisy conditions compared with SIFT and repeated SIFT.  We expect this increased performance applies to LFs collected in low light, and also to shadowed and low-contrast regions of well-lit scenes. It also applies where contrast is limited by participating media like water, dust, smoke, or fog.

\setlength{\tabcolsep}{1pt}
\begin{table*}
%\relsize{-1.5}
\begin{center}
%\begin{tabular}{p{4em}|p{4em}|p{4em}|p{4.5em}|p{4.5em}|p{4em}|p{4.5em}|p{4em}|p{3.5em}|p{4em}}
\begin{tabular}{p{7em}p{5em}p{5em}p{5.5em}p{5.5em}p{4.5em}p{5.5em}p{5em}p{5em}p{4em}}
\toprule
Method 	& \% pass 	& Keypts / Img	& Putative Matches / Img& Inlier Matches / Img	& Match Ratio& Precision& Matching Score& 3D Points& Track Len\\
%\midrule 
%Results as presented at final Intel VEC retreat
%Defaults\\
%LiFF 	& \textbf{64.19}		& 2707	& \textbf{218}	& \textbf{210.89}	&\textbf{0.11}	&\textbf{0.93}	&\textbf{0.11}	&\textbf{245.23}	&\textbf{2.17} \\
%SIFT 	& 57.83		& 2698	&176	&169.18		&0.08	&0.92	&0.07	&195.07	&1.91 \\
%\midrule
%Permissive\\
%LiFF 	& 97.53 	& 2707 	& \textbf{218}	& \textbf{210.91}	& \textbf{0.11}	& \textbf{0.93}	& \textbf{0.11}	& \textbf{459.86}	& \textbf{2.40} \\
%SIFT 	& \textbf{97.88} 	& 2698	& 176	& 169.16	& 0.08	& 0.92	& 0.07	& 387.23	& 2.35 \\
%Results as presented at Intel;  
\midrule
% Results after re-running with latest colmap performance
\multicolumn{2}{l}{COLMAP Defaults}\\
LiFF	& \textbf{64.19}	& 2684	& \textbf{282}	& \textbf{274}	& \textbf{0.14}	& \textbf{0.96}	& \textbf{0.13}	& \textbf{382}	& \textbf{3.38}\\
SIFT	& 57.83	& 2669	& 243	& 235	& 0.10	& 0.95	& 0.10	& 337	& 3.31\\
\midrule
\multicolumn{2}{l}{COLMAP Permissive}\\
LiFF	& 97.53	& 2689	& \textbf{213}	& \textbf{206}	& \textbf{0.11}	&  \textbf{0.93}	& \textbf{0.11}	&  \textbf{472}	& \textbf{2.46}\\
SIFT	& \textbf{97.88}	& 2688	& 175	& 167	& 0.077	& 0.92	& 0.073	& 396	& 2.40\\
\midrule
\multicolumn{2}{l}{Defaults Intersect}\\
LiFF	&  54.65	&  2674	&  \textbf{304}	&  \textbf{297}	&  \textbf{0.15}	&  \textbf{0.96}	&  \textbf{0.14} &  \textbf{418}	&  \textbf{3.44}\\
SIFT	&  54.65	&  2689	&  248	&  240	&  0.10	&  0.95	&  0.10	&  348	&  3.33\\ 
\midrule
\multicolumn{2}{l}{Permissive Intersect}\\
LiFF	&  96.23	&  2687	&  \textbf{212}	&  \textbf{205}	&  \textbf{0.11}	 & \textbf{0.93}	&  \textbf{0.11}	&   \textbf{473}	&  \textbf{2.46}\\  
SIFT	&  96.23	&  2684	&  172	&  165	&  0.076 & 0.92	&  0.073 &  397	&  2.40\\
\bottomrule
\end{tabular}
\end{center}
\caption{
\protect\begin{flushleft}
Structure-from-motion: With COLMAP's default values, LiFF outperforms SIFT in all measures, including successful reconstruction of significantly more scenes; with more permissive settings, COLMAP reconstructs nearly all scenes, succeeding on slightly more scenes using SIFT, but with LiFF outperforming SIFT in all other measures including 3D points per model. Taking only those scenes that passed with both feature detectors (''Intersect'') allows a direct comparison of performance, with LiFF outperforming SIFT in all cases.
\protect\end{flushleft}
}
\label{Tab_SfM}
\end{table*}

%---
\subsection{Structure from Motion}

%---
Following the feature comparison approach in~\cite{schonberger2017comparative}, we employed an SfM solution to evaluate LiFF in the context of 3D reconstruction applications.  We used a Lytro Illum to collect a large dataset of \glspl{LF} with multiple views of each scene.  The dataset contains 4211 \glspl{LF} covering 850 scenes in 30 categories, with between 3 and 5 views of each scene.  Images are in indoor and outdoor campus environments, and include examples of Lambertian and non-Lambertian surfaces, occlusion, specularity, subsurface scattering, fine detail, and transparency.  No attempt was made to emphasize challenging content.

Although we expect LiFF's slope estimates could dramatically improve SfM, we ignore this information to allow a more direct comparison with SIFT.  We also use identical settings for all parameters common to SIFT and LiFF.  Based on the noise performance experiments above, a higher peak threshold for LiFF would likely result in fewer spurious features without loss of useful features.  However, by using identical thresholds we are better able to highlight the behavioural differences between LiFF and SIFT, rather than focusing exclusively on the difference in noise performance.

We extracted the central view of each \gls{LF} and converted to grayscale. The method of grayscale conversion significantly impacts performance~\cite{kanan2012color}, and we determined that MATLAB's luminance-based conversion followed by gamma correction with a factor of 0.5, and finally histogram equalization, yielded good results.

We ran LiFF and the VLFeat implementation of SIFT using a peak threshold of 0.0066, edge threshold 10, and DoG scales covering 4 octaves over 3 levels per octave.  We started at octave -1 because our images are relatively small, making smaller features important.  For LiFF we employed the centermost $11 \times 11$ subimages, and computed the focal stack over 11 slopes between -1 and 1.

For the feature descriptor we found that L1 root normalization yields significantly improved matching compared with the default L2 normalization build into VLFeat's implementation of SIFT.  We therefore applied this same normalization scheme to both SIFT and LiFF feature descriptors.  To confirm that our external feature detection was working correctly, we compared COLMAP's performance when using our externally extracted SIFT features and when using its internal calls to SIFT, and achieved virtually identical results.

We ran COLMAP up to and including the SfM stage, stopping before dense multi-view stereo reconstruction.  We evaluated performance in terms of numbers of keypoints per image, putative feature matches generated per image, and number of putative matches classified as inliers during SfM. Following~\cite{schonberger2017comparative}, we also evaluated the putative match ratio: the proportion of detected features that yield putative matches; precision: the proportion of putative matches yielding inlier matches; matching score: the proportion of features yielding inlier matches; the mean number of 3D points in the reconstructed models; and track length: the mean number of images over which a feature is successfully tracked.

With its default settings, we found that COLMAP failed to generate output for many scenes.  It failed to converge during bundle adjustment, or failed to identify a good initial image pair.  With each of our images having only $541 \times 376$ pixels, and each scene only 3 to 5 images, COLMAP's default settings are not well suited to our dataset.  The difference in performance between LiFF and SIFT at this stage is nevertheless informative, and is shown in the top row of Table~\ref{Tab_SfM}.  LiFF did not detect many more features than SIFT, but it did result in a significantly higher number of successfully reconstructed scenes (\% pass).  The statistics support the conclusion that LiFF has a higher proportion of informative features, yielding higher absolute numbers of putative and inlier matches, higher proportions of inlier matches, more 3D points, and longer track lengths.  Note that we have not highlighted the higher keypoint count as being a superior result, as having LiFF detect more features is not necessarily a better outcome without those features also being useful.

We relaxed COLMAP's settings to better deal with our dataset, reducing the mapper's minimum inlier counts, minimum track length, and minimum 3D point count.  In this more permissive mode COLMAP was able to reconstruct most of the scenes in the dataset.  As seen in the second set of results in Table.~\ref{Tab_SfM}, in this mode SIFT allowed slightly more scenes to be reconstructed, and detected a nearly identical number of features, but performed dramatically less well than LiFF in all other statistics.  Note in particular that LiFF-generated models had on average 472 reconstructed points compared with SIFT's 396.

A shortcoming of the comparisons made above is that they are applied over different subsets of the data: SIFT passed a different set of scenes than LiFF.  For a fair comparison we computed the same statistics over only those scenes that passed using both SIFT and LiFF features.  The results, in the bottom half of Table~\ref{Tab_SfM}, clearly show LiFF outperforming SIFT in all measures.

\begin{figure}
	\centering
	\includegraphics[width=\hsize]{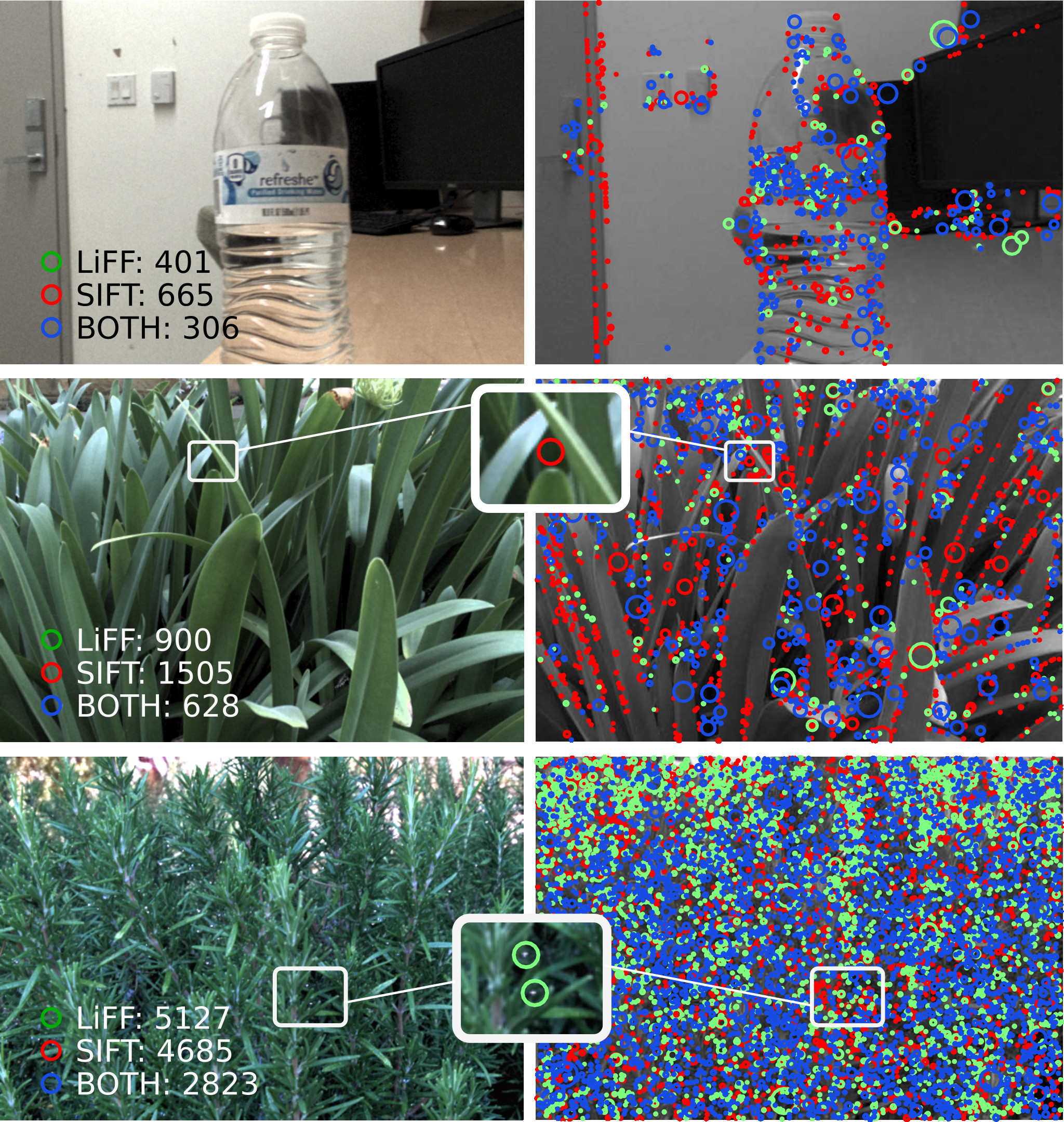}\hfil
	\caption[]{Comparison to SIFT: Features identified only by LiFF, only by SIFT, and by both are shown in green, red, and blue respectively.  (top) LiFF rejects spurious features in low-contrast areas and to some extent those distorted through refraction; (center) LiFF rejects spurious features at occlusion boundaries -- the inset highlights a SIFT-only detection caused by leaves at different depths; (bottom) LiFF detects partially occluded features missed by SIFT -- note the increasing proportion of LiFF-only features toward the back of the scene, and the LiFF-only detections highlighted in the inset. Slope estimates for the bottom scene are shown in Fig.~\ref{Fig_3D}.%
%\vspace{-0.5em}
}
\label{Fig_ChallengingCases}
\end{figure}

%---
\subsection{Challenging Cases}

To better expose the differences in performance between SIFT and LiFF, we investigated those scenes for which COLMAP had trouble converging with SIFT features, but passed when using LiFF features.  Fig.~\ref{Fig_ChallengingCases} depicts some informative examples.  At right we show features detected only by LiFF (green), only by SIFT (red), and by both methods (blue).  In the top row we see that this relatively well-lit indoor scene has low contrast around the door edge yielding many spurious SIFT-only detections.  Note also that the texture refracted through the water bottle triggers some SIFT-only detections. The inconsistent apparent motion of refracted features make them undesirable for SfM, and the lack of a well-defined depth prevents LiFF from detecting these as features.

The center row in Fig.~\ref{Fig_ChallengingCases} shows a scene with many spurious SIFT detections near edges, but also at occlusion boundaries.  SIFT cannot distinguish between well-defined shapes and those formed by the chance alignment of occluding objects.  LIFF on the other hand rejects shapes formed by occluding objects at different depths, because these do not have a well-defined depth.  An example of a spurious occlusion feature detected only by SIFT is highlighted in the inset.

\begin{figure}
	\centering
	\includegraphics[width=\hsize]{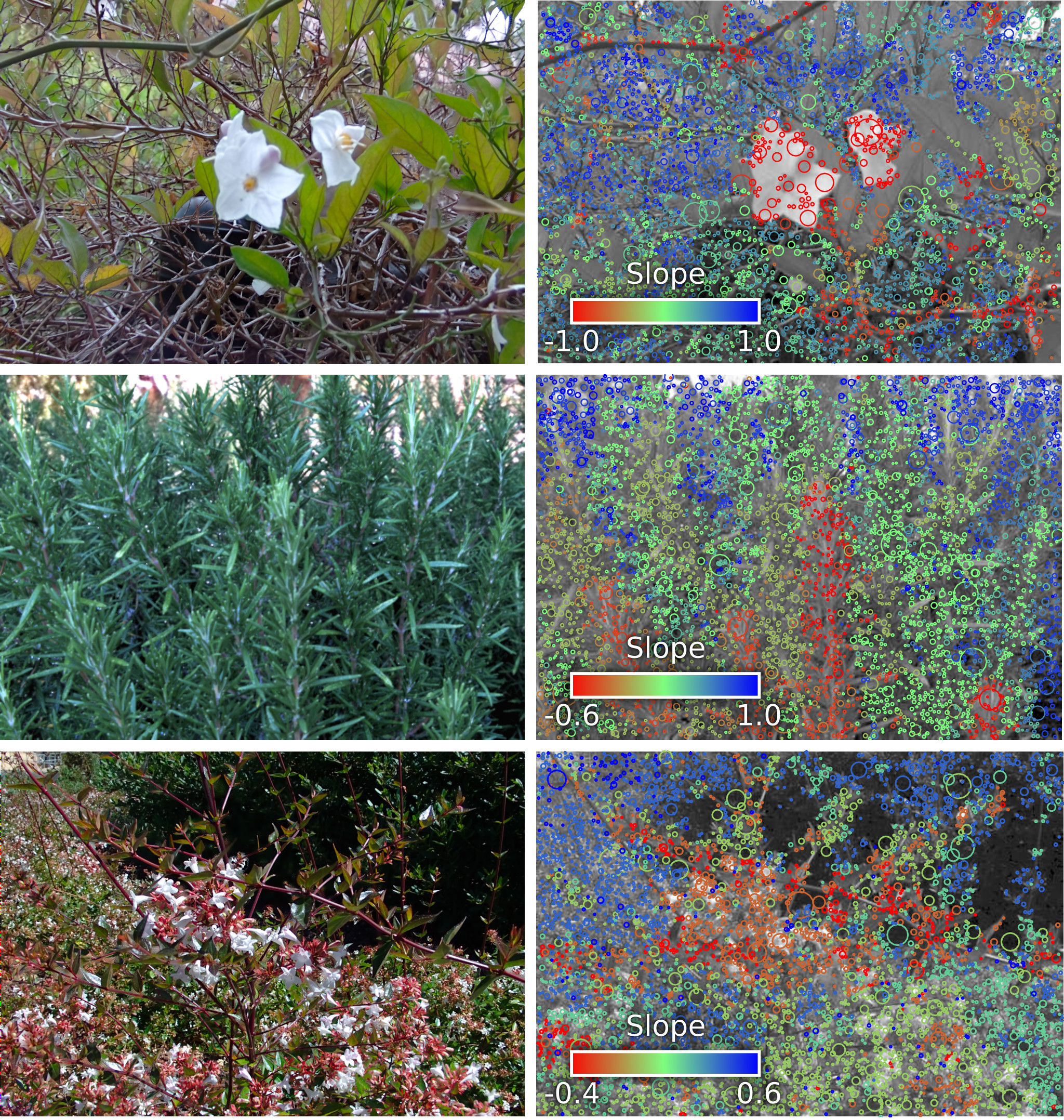}
	\caption[]{3D Scene Shape: In this work we establish LiFF's ability to deliver more informative features by virtue of higher selectivity and ability to image through partial occlusion.  We expect LiFF's slope estimates will also be of substantial interest.  Here we see the 3D shape of each scene revealed through the slopes of the detected LiFF features.%
%\vspace{-1em}
}
\label{Fig_3D}
\end{figure}

The bottom row in Fig.~\ref{Fig_ChallengingCases} shows a scene for which LiFF delivers more features than SIFT.  Notice the increasing proportion of LiFF-only features towards the back of the scene, where most of the features are partially occluded by foreground elements.  In the inset we see an example of two water droplets just visible through foreground occlusions, detected only by LiFF. In more extreme cases, features may be entirely blocked in some subimages but still visible to LiFF.  Note that the green circles in the inset are expanded to aid clarity.  This scene is repeated in Fig.~\ref{Fig_3D}, which provides visual confirmation that 3D structure is being reflected in the LiFF slope estimates.

%--------------------------------------------------------
\section{Conclusion}

We presented LiFF, a feature detector and descriptor for \glspl{LF} that directly extends SIFT to operate on the entire \gls{LF}.  The proposed detector is faster than the common practice of repeating SIFT across multiple views, and produces more correct detections and fewer spurious detections in challenging conditions.  We demonstrate an $18 \times$ speed increase on Lytro Illum-captured imagery compared with repeated SIFT, and anticipate further optimization is possible via parallelization and implementation on GPU.

In SfM tests, we showed LiFF to outperform SIFT in terms of absolute numbers of putative and inlier matches, proportions of inlier matches,  numbers of images through which features are tracked, and  the numbers of 3D points in the reconstructed models.  Our test dataset was not manipulated to emphasize challenging scenes, these results are for typical indoor and outdoor environments.  We expect that in more challenging conditions LiFF can even more dramatically improve the performance of 3D reconstruction, and expand the range of applications in which feature-based techniques can be applied.

As future work we expect that adaptive selection of focal stack slopes could further improve the speed of LiFF. An interesting benefit of the focal stack is that it can be trivially extended to perform linear super-resolution~\cite{ihrke2016principles}, allowing finer features to be detected, though at the cost of increased processing time. 

Recent work has shown that computing histograms over multiple scales offers improved SIFT detector performance, and this can also be applied to LiFF features~\cite{dong2015domain, schonberger2017comparative}.  We also anticipate the slope information that LiFF recovers to be of interest.  For a calibrated \gls{LF} camera, slope yields an absolute 3D position and absolute scale for each feature.  This absolute scale can be employed as a discriminator in a scale-\emph{sensitive} approach to feature matching. Finally, the 3D information retrieved by LiFF may be of significant utility in directly informing 3D reconstruction.

\textbf{Acknowledgments} 
This work was supported in part by the NSF/Intel Partnership on Visual and Experiential Computing (Intel \#1539120, NSF \#IIS-1539120).

%-------------------------------------------------------------------------
{\small
\bibliographystyle{ieee}
\bibliography{LightFieldBib}

\begin{thebibliography}{10}\itemsep=-1pt

\bibitem{adelson1992single}
E.~H. Adelson and J.~Y.~A. Wang.
\newblock Single lens stereo with a plenoptic camera.
\newblock {\em IEEE Transactions on Pattern Analysis and Machine Intelligence
  ({TPAMI})}, 14(2):99--106, 1992.

\bibitem{bay2008surf}
H.~Bay, A.~Ess, T.~Tuytelaars, and L.~V. Gool.
\newblock Speeded-up robust features ({SURF}).
\newblock {\em Computer Vision and image understanding}, 110(3):346--359, 2008.

\bibitem{bishop2009light}
T.~E. Bishop, S.~Zanetti, and P.~Favaro.
\newblock Light field superresolution.
\newblock In {\em Computational Photography ({ICCP})}, pages 1--9. IEEE, 2009.

\bibitem{bok2014geometric}
Y.~Bok, H.-G. Jeon, and I.~S. Kweon.
\newblock Geometric calibration of micro-lens-based light-field cameras using
  line features.
\newblock In {\em Computer Vision--ECCV 2014}, pages 47--61. Springer, 2014.

\bibitem{bolles1987epipolar}
R.~Bolles, H.~Baker, and D.~Marimont.
\newblock Epipolar-plane image analysis: An approach to determining structure
  from motion.
\newblock {\em Intl. Journal of Computer Vision ({IJCV})}, 1(1):7--55, 1987.

\bibitem{dansereau2013decoding}
D.~G. Dansereau, O.~Pizarro, and S.~B. Williams.
\newblock Decoding, calibration and rectification for lenselet-based plenoptic
  cameras.
\newblock In {\em Computer Vision and Pattern Recognition ({CVPR})}, pages
  1027--1034. IEEE, June 2013.

\bibitem{dansereau2015linear_paper}
D.~G. Dansereau, O.~Pizarro, and S.~B. Williams.
\newblock Linear volumetric focus for light field cameras.
\newblock {\em ACM Transactions on Graphics (TOG)}, 34(2):15, Feb. 2015.

\bibitem{dansereau2017wide}
D.~G. Dansereau, G.~Schuster, J.~Ford, and G.~Wetzstein.
\newblock A wide-field-of-view monocentric light field camera.
\newblock In {\em Computer Vision and Pattern Recognition ({CVPR})}, pages
  3757--3766. IEEE, July 2017.

\bibitem{davis2012unstructured}
A.~Davis, M.~Levoy, and F.~Durand.
\newblock Unstructured light fields.
\newblock In {\em Computer Graphics Forum}, volume~31, pages 305--314. Wiley
  Online Library, 2012.

\bibitem{dong2013plenoptic}
F.~Dong, S.-H. Ieng, X.~Savatier, R.~Etienne-Cummings, and R.~Benosman.
\newblock Plenoptic cameras in real-time robotics.
\newblock {\em Intl. Journal of Robotics Research ({IJRR})}, 32(2):206--217,
  2013.

\bibitem{dong2015domain}
J.~Dong and S.~Soatto.
\newblock Domain-size pooling in local descriptors: Dsp-sift.
\newblock In {\em Computer Vision and Pattern Recognition ({CVPR})}, pages
  5097--5106, 2015.

\bibitem{gao2015robust}
X.~Gao and T.~Zhang.
\newblock Robust {RGB-D} simultaneous localization and mapping using planar
  point features.
\newblock {\em Robotics and Autonomous Systems}, 72:1--14, 2015.

\bibitem{georgiev2008unified}
T.~Georgiev, C.~Intwala, S.~Babakan, and A.~Lumsdaine.
\newblock Unified frequency domain analysis of lightfield cameras.
\newblock In {\em Computer Vision--ECCV 2008}, pages 224--237. Springer, 2008.

\bibitem{ghasemi2014scale}
A.~Ghasemi and M.~Vetterli.
\newblock Scale-invariant representation of light field images for object
  recognition and tracking.
\newblock In {\em Proceedings of the SPIE}, volume 9020. Intl. Society for
  Optics and Photonics, 2014.

\bibitem{gortler1996lumigraph}
S.~Gortler, R.~Grzeszczuk, R.~Szeliski, and M.~Cohen.
\newblock The lumigraph.
\newblock In {\em {SIGGRAPH}}, pages 43--54. ACM, 1996.

\bibitem{gumhold2001feature}
S.~Gumhold, X.~Wang, and R.~S. MacLeod.
\newblock Feature extraction from point clouds.
\newblock In {\em IMR}, 2001.

\bibitem{gupta2014learning}
S.~Gupta, R.~Girshick, P.~Arbel{\'a}ez, and J.~Malik.
\newblock Learning rich features from {RGB-D} images for object detection and
  segmentation: Supplementary material, 2014.

\bibitem{hanrahan2006digital}
P.~Hanrahan and R.~Ng.
\newblock Digital correction of lens aberrations in light field photography.
\newblock In {\em Intl. Optical Design Conference}, page WB2. Optical Society
  of America, 2006.

\bibitem{ihrke2016principles}
I.~Ihrke, J.~Restrepo, and L.~Mignard-Debise.
\newblock Principles of light field imaging.
\newblock {\em IEEE Signal Processing Magazine}, 1053(5888/16), 2016.

\bibitem{johannsen2015linear}
O.~Johannsen, A.~Sulc, and B.~Goldluecke.
\newblock On linear structure from motion for light field cameras.
\newblock In {\em Intl. Conference on Computer Vision ({ICCV})}, pages
  720--728, 2015.

\bibitem{kanan2012color}
C.~Kanan and G.~W. Cottrell.
\newblock Color-to-grayscale: does the method matter in image recognition?
\newblock {\em PloS one}, 7(1):e29740, 2012.

\bibitem{levin2010linear}
A.~Levin and F.~Durand.
\newblock Linear view synthesis using a dimensionality gap light field prior.
\newblock In {\em Computer Vision and Pattern Recognition ({CVPR})}, pages
  1831--1838. IEEE, 2010.

\bibitem{levoy1996light}
M.~Levoy and P.~Hanrahan.
\newblock Light field rendering.
\newblock In {\em {SIGGRAPH}}, pages 31--42. ACM, 1996.

\bibitem{liang2015light}
C.-K. Liang and R.~Ramamoorthi.
\newblock A light transport framework for lenslet light field cameras.
\newblock {\em ACM Transactions on Graphics ({TOG})}, 34(2):16, 2015.

\bibitem{lowe2004distinctive}
D.~G. Lowe.
\newblock Distinctive image features from scale-invariant keypoints.
\newblock {\em Intl. Journal of Computer Vision ({IJCV})}, 60(2):91--110, 2004.

\bibitem{maeno2013light}
K.~Maeno, H.~Nagahara, A.~Shimada, and R.~I. Taniguchi.
\newblock Light field distortion feature for transparent object recognition.
\newblock In {\em Computer Vision and Pattern Recognition ({CVPR})}, pages
  2786--2793. IEEE, June 2013.

\bibitem{marichal2011fast}
J.~G. Marichal-Hern{\'a}ndez, J.~P. L{\"u}ke, F.~L. Rosa, and J.~M.
  Rodr{\'\i}guez-Ramos.
\newblock Fast approximate {4D}: {3D} discrete radon transform, from light
  field to focal stack with o(n4) sums.
\newblock In {\em IS\&T/SPIE Electronic Imaging}, pages 78710G--78710G. Intl.
  Society for Optics and Photonics, 2011.

\bibitem{marwah2013compressive}
K.~Marwah, G.~Wetzstein, Y.~Bando, and R.~Raskar.
\newblock Compressive light field photography using overcomplete dictionaries
  and optimized projections.
\newblock In {\em {SIGGRAPH}}, volume~32, pages 1--11, New York, NY, USA, 2013.
  ACM.

\bibitem{mitra2014framework}
K.~Mitra, O.~S. Cossairt, and A.~Veeraraghavan.
\newblock A framework for analysis of computational imaging systems: role of
  signal prior, sensor noise and multiplexing.
\newblock {\em IEEE Transactions on Pattern Analysis and Machine Intelligence
  ({TPAMI})}, 36(10):1909--1921, 2014.

\bibitem{ng2005fourier}
R.~Ng.
\newblock Fourier slice photography.
\newblock {\em ACM Transactions on Graphics ({TOG})}, 24(3):735--744, July
  2005.

\bibitem{ng2005light}
R.~Ng, M.~Levoy, M.~Br{\'e}dif, G.~Duval, M.~Horowitz, and P.~Hanrahan.
\newblock Light field photography with a hand-held plenoptic camera.
\newblock Technical report, Stanford University Computer Science, 2005.

\bibitem{perez2014fast}
F.~P{\'e}rez, A.~P{\'e}rez, M.~Rodr{\'\i}guez, and E.~Magdaleno.
\newblock A fast and memory-efficient discrete focal stack transform for
  plenoptic sensors.
\newblock {\em Digital Signal Processing}, 2014.

\bibitem{rosten2006machine}
E.~Rosten and T.~Drummond.
\newblock Machine learning for high-speed corner detection.
\newblock In {\em European Conference on Computer Vision ({ECCV})}, pages
  430--443. Springer, 2006.

\bibitem{rublee2011orb}
E.~Rublee, V.~Rabaud, K.~Konolige, and G.~Bradski.
\newblock {ORB}: An efficient alternative to {SIFT} or {SURF}.
\newblock In {\em Intl. Conference on Computer Vision ({ICCV})}, pages
  2564--2571. IEEE, 2011.

\bibitem{schonberger2016structure}
J.~L. Schonberger and J.-M. Frahm.
\newblock Structure-from-motion revisited.
\newblock In {\em Proceedings of the IEEE Conference on Computer Vision and
  Pattern Recognition}, pages 4104--4113, 2016.

\bibitem{schonberger2017comparative}
J.~L. Sch{\"o}nberger, H.~Hardmeier, T.~Sattler, and M.~Pollefeys.
\newblock Comparative evaluation of hand-crafted and learned local features.
\newblock In {\em Computer Vision and Pattern Recognition ({CVPR})}, July 2017.

\bibitem{smith2009light}
B.~Smith, L.~Zhang, H.~Jin, and A.~Agarwala.
\newblock Light field video stabilization.
\newblock In {\em Intl. Conference on Computer Vision ({ICCV})}, pages
  341--348. IEEE, 2009.

\bibitem{srinivasan2017light}
P.~P. Srinivasan, R.~Ng, and R.~Ramamoorthi.
\newblock Light field blind motion deblurring.
\newblock In {\em Computer Vision and Pattern Recognition ({CVPR})}, 2017.

\bibitem{srinivasan2017learning}
P.~P. Srinivasan, T.~Wang, A.~Sreelal, R.~Ramamoorthi, and R.~Ng.
\newblock Learning to synthesize a {4D} {RGBD} light field from a single image.
\newblock In {\em Intl. Conference on Computer Vision ({ICCV})}, pages
  2262--2270, 2017.

\bibitem{steder2010narf}
B.~Steder, R.~B. Rusu, K.~Konolige, and W.~Burgard.
\newblock {NARF}: {3D} range image features for object recognition.
\newblock In {\em Intelligent Robots and Systems (IROS) Workshops}, volume~44,
  2010.

\bibitem{tao2015depth}
M.~W. Tao, P.~P. Srinivasan, J.~Malik, S.~Rusinkiewicz, and R.~Ramamoorthi.
\newblock Depth from shading, defocus, and correspondence using light-field
  angular coherence.
\newblock In {\em Computer Vision and Pattern Recognition ({CVPR})}, pages
  1940--1948, 2015.

\bibitem{teixeira2017epipolar}
J.~A. Teixeira, C.~Brites, F.~Pereira, and J.~Ascenso.
\newblock Epipolar based light field key-location detector.
\newblock In {\em Multimedia Signal Processing (MMSP)}, pages 1--6. IEEE, 2017.

\bibitem{tian2017depth}
J.~Tian, Z.~Murez, T.~Cui, Z.~Zhang, D.~Kriegman, and R.~Ramamoorthi.
\newblock Depth and image restoration from light field in a scattering medium.
\newblock In {\em Intl. Conference on Computer Vision ({ICCV})}, 2017.

\bibitem{tosic20143d}
I.~To{\v{s}}i{\'c} and K.~Berkner.
\newblock {3D} keypoint detection by light field scale-depth space analysis.
\newblock In {\em Image Processing ({ICIP})}, pages 1927--1931. IEEE, 2014.

\bibitem{veeraraghavan2007dappled}
A.~Veeraraghavan, R.~Raskar, A.~Agrawal, A.~Mohan, and J.~Tumblin.
\newblock Dappled photography: Mask enhanced cameras for heterodyned light
  fields and coded aperture refocusing.
\newblock {\em ACM Transactions on Graphics ({TOG})}, 26(3):69, 2007.

\bibitem{wang2017light}
T.-C. Wang, J.-Y. Zhu, N.~K. Kalantari, A.~A. Efros, and R.~Ramamoorthi.
\newblock Light field video capture using a learning-based hybrid imaging
  system.
\newblock {\em ACM Transactions on Graphics (Proceedings of SIGGRAPH 2017)},
  36(4), 2017.

\bibitem{wanner2012globally}
S.~Wanner and B.~Goldluecke.
\newblock Globally consistent depth labeling of {4D} light fields.
\newblock In {\em Computer Vision and Pattern Recognition ({CVPR})}, pages
  41--48. IEEE, 2012.

\bibitem{wanner2013reconstructing}
S.~Wanner and B.~Goldluecke.
\newblock Reconstructing reflective and transparent surfaces from epipolar
  plane images.
\newblock In {\em German Conference on Pattern Recognition}, pages 1--10.
  Springer, 2013.

\bibitem{wanner2013globally}
S.~Wanner, C.~Straehle, and B.~Goldluecke.
\newblock Globally consistent multi-label assignment on the ray space of {4D}
  light fields.
\newblock In {\em Computer Vision and Pattern Recognition ({CVPR})}. IEEE, June
  2013.

\bibitem{wetzstein2011computational}
G.~Wetzstein, I.~Ihrke, D.~Lanman, and W.~Heidrich.
\newblock Computational plenoptic imaging.
\newblock In {\em Computer Graphics Forum}, volume~30, pages 2397--2426. Wiley
  Online Library, 2011.

\bibitem{wilburn2005high}
B.~Wilburn, N.~Joshi, V.~Vaish, E.~Talvala, E.~Antunez, A.~Barth, A.~Adams,
  M.~Horowitz, and M.~Levoy.
\newblock High performance imaging using large camera arrays.
\newblock {\em ACM Transactions on Graphics ({TOG})}, 24(3):765--776, 2005.

\bibitem{xu2015camera}
Y.~Xu, K.~Maeno, H.~Magahara, and R.~I. Taniguchi.
\newblock Camera array calibration for light field acquisition.
\newblock {\em Frontiers of Computer Science}, 9(5):691--702, 2015.

\bibitem{xu2015transcut}
Y.~Xu, H.~Nagahara, A.~Shimada, and R.~I. Taniguchi.
\newblock Transcut: transparent object segmentation from a light-field image.
\newblock In {\em Intl. Conference on Computer Vision ({ICCV})}, pages
  3442--3450, 2015.

\bibitem{zhang2017ray}
Y.~Zhang, P.~Yu, W.~Yang, Y.~Ma, and J.~Yu.
\newblock Ray space features for plenoptic structure-from-motion.
\newblock In {\em Computer Vision and Pattern Recognition ({CVPR})}, pages
  4631--4639, 2017.

\bibitem{zhong2009intrinsic}
Y.~Zhong.
\newblock Intrinsic shape signatures: A shape descriptor for {3D} object
  recognition.
\newblock In {\em Computer Vision (ICCV) Workshops}, pages 689--696. IEEE,
  2009.

\end{thebibliography}
}

\end{document}